\documentclass[a4paper,UKenglish,cleveref, autoref, thm-restate,authorcolumns]{lipics-v2019}

\usepackage{gensymb}

\title{Geo-Situation for Modeling Causality of Geo-Events in Knowledge Graphs} 

\titlerunning{Geo-Situation for Modeling Causality of Geo-Events in Knowledge Graphs} 

\author{Shirly Stephen}{School of Geographic Sciences and Urban Planning, Arizona State University, Tempe, AZ, USA}{}{}{}{}{}

\author{Wenwen Li}{School of Geographic Sciences and Urban Planning, Arizona State University, Tempe, AZ, USA}{}{}{}
\author{Torsten Hahmann}{School of Computing and Information Sciences, University of Maine, Orono, ME, USA}{}{}{}

\authorrunning{S.\, Stephen and W.\, Li and T.\, Hahmann} 

\Copyright{Shirly Stephen and Wenwen Li and Torsten Hahmann} 


\category{} 

\relatedversion{} 

\acknowledgements{This work is in part supported by the National Science Foundation under Grants OIA-2033521, BCS-1853864, BCS-1455349, and GCR-2021147.}

\nolinenumbers 

\begin{document}

\maketitle

\begin{abstract}

\noindent This paper proposes a framework for representing and reasoning causality between geographic events by introducing the notion of \textit{Geo-Situation}. This concept links to observational snapshots that represent sets of conditions, and either acts as the setting of a geo-event or influences the initiation of a geo-event. We envision the use of this framework within knowledge graphs that represent geographic entities will help answer the important question of why a geographic event occurred.
\end{abstract}
\setlength{\jot}{10pt} 

\newcommand{\satisfies}{{\mathit{satisfies}}}
\newcommand{\situated}{{\mathit{situated}}}

\newcommand{\ofEntity}{{\mathit{ofEntity}}}
\newcommand{\hasMeasurement}{{\mathit{hasMeasurement}}}
\newcommand{\extent}{{\mathit{extent}}}
\newcommand{\effects}{{\mathit{effects}}}
\newcommand{\affects}{{\mathit{affects}}}
\newcommand{\PC}{{\mathit{PC}}}
\newcommand{\causes}{{\mathit{causes}}}
\newcommand{\WindSpeed}{{\mathit{WindSpeed}}}
\newcommand{\SeaSurfaceTemp}{{\mathit{SeaSurfaceTemp}}}

\newcommand{\GeoEvent}{\mathit{Geo}\text{-}\mathit{Event}}
\newcommand{\GeoObject}{\mathit{Geo}\text{-}\mathit{Object}}
\newcommand{\GeoSituation}{\mathit{Geo}\text{-}\mathit{Situation}}
\newcommand{\TemporalEntity}{\mathit{Temporal}\text{-}\mathit{Entity}}
\newcommand{\TropicalStorm}{\mathit{Tropical}\text{-}\mathit{Storm10}}
\newcommand{\TropicalStormn}{\mathit{Tropical}\text{-}\mathit{Storm12}}
\newcommand{\HurricaneKatrina}{\mathit{Hurricane}\text{-}\mathit{Katrina}}

\vspace{-0.2cm}
\section{Introduction}\label{sec:Introduction}
\vspace{-0.3cm}
The Thomas wildfire, one of the largest recent wildfires in California destroyed at least 1,063 structures, but also caused secondary damage by depositing ash on coastal and agricultural lands of Southern California. While the fire was initiated when power lines collided during a high wind event, during the fire event, unfavorable conditions caused by temperatures, relative humidity and wind speed magnified the subsequent event. Similarly, Hurricane Katrina consisted of 23 distinct spatio-temporally co-occuring events (such as HeavyRain, FlashFlood, DebrisFlow, TropicalStorm) as recorded in the NOAA (National Oceanic and Atmospheric Administration) storm database. Each event in turn caused sub-events such as property damage and human fatalities. Most of these events are causally related, while some are effected when a prescribed set of conditions are satisfied. Modeling this high-level causal knowledge is essential (1) to represent many event-datasets meaningfully as RDF triples in a knowledge graph, (2) to understand and reason about these events and their interaction with the environment, infrastructure and physical nature, (3) to facilitate deduction from events and their full extent of related losses.

Existing process- and event-centric geographic models can answer questions of when, where and what happened. But they cannot answer \emph{why}, specifically they cannot reveal the underlying environmental factors that abet these events. Causalities have by large been considered a part of geospatial analytics rather than as a knowledge modeling task. Determining causes often requires expert knowledge, which can be encoded as an ontology to help detect and learn event causation from data. In this paper, we propose a geo-event ontological model that (1) is interoperable by extending concepts from a foundational ontology and other OGC, W3C standards, (2) builds on existing formal notions of geographic endurants and perdurants to model geographic-event datasets, (3) introduces the notion of \textit{Geo-Situation} that describes the setting of geographic objects and events in terms of observational attributes and their measured values, (4) enables finding patterns and interdependent knowledge based on (geographic) object-event relationships, their settings and event causality.

\vspace{-0.3cm}
\section{Related Work} 
\vspace{-0.3cm}

\noindent \textbf{General Ontologies of Events and Causation: }
Upper level ontologies such as DOLCE (Descriptive Ontology for Linguistic and Cognitive Engineering) and Basic Formal Ontology distinguish an \textit{Endurant} that persists through time from a \textit{Perdurant} that unfolds through time, and which are related through the general \textit{participation} relation. Perdurants are further refined based on temporal scale, properties of individual parts and material participants into \textit{event}, \textit{state}, \textit{process}. Except for states, other perdurants are causally effectual. Galton goes a step further to define 6 causal relations \cite{galton2012states} by restricting their domain and range based on their participant's role. Dedicated event models either use actions such as in Event-Calculus \cite{shanahan1999event}, or domain and lexical pattern descriptions in Event-Model-F \cite{scherp2012core} to represent causality.  However none of them model the role played by environmental attributes within a geographic setting, which is key to explaining and extracting causality between geographical events.

\noindent \textbf{Causal Modeling for Geographic Perdurants:} 
Most process-centric \cite{grenon2004snap, stephen2017ontological} and event-centric \cite{zhang2005hierarchies,worboys2005event} geographic models describe the environment from three perspectives -- (geographic-) processes, events, and elements. They also express their spatio-temporal interactions, and describe the evolution of elements. Some \cite{worboys2005event} model causality similar to general event models. \cite{zhang2005hierarchies} uses the notion of precondition (by comparing properties of objects at different times) to describe the occurrence an event. But oftentimes it is not just an object's property but a more complex geographic situation that instigates a geographic event (e.g. Fig.~\ref{example}). \cite{devaraju2015formal} adopts the notion of process that actuates sensors to make observations, which are then used to infer events, and \cite{llaves2014event} describes variations in quantitative observed properties as events. However, most geographic event-datasets do not contain process-related data, and therefore building a knowledge graph that complies with this model is unrealistic. For example, NOAA's storm dataset does not record observations for every sub-event of a larger geo-event. In the absence of observations, asserting the sequence of causality between individual parts of a larger event is impractical. 
We therefore adopt an approach that models causation of an event based on observations within a larger associated situation, \emph{and/or} due to another event.

\vspace{-0.4cm}
\section{A Schema-Level Formalism for Causal Relations for Geo-Events}\label{methodology}
\begin{figure}[h]
\caption{A schema of basic concepts for modeling and inferring causation for geographic events. Relations in the Ge-Events schema are prefixed with the ontology they are inherited from.}
\centering
\includegraphics[scale=0.5]{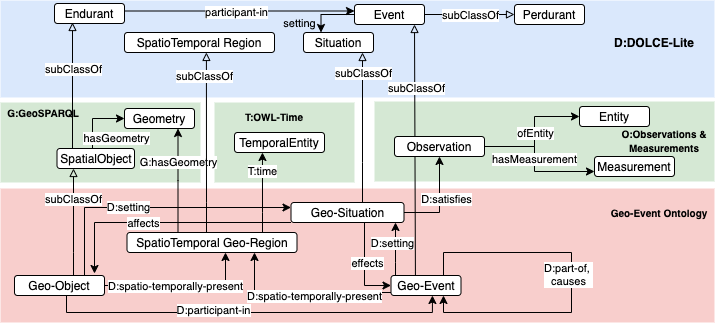}
\label{schema}
\end{figure}

\vspace{-0.3cm}
The geo-event ontological model that we propose (Fig.~\ref{schema}) is constructed as an extension of the foundational ontology DOLCE (we use the OWL version \href{https://www.w3.org/2001/sw/BestPractices/WNET/DLP3941_daml.html}{{\color{blue}DOLCE-Lite}}), GeoSPARQL, OWL-Time and Observations and Measurements (O$\&$M). To define the functional requirements of the model, we use \href{https://www.ncdc.noaa.gov/stormevents/}{{\color{blue}NOAA's storm dataset}} dataset as a use-case. The key concepts and relations are described below.

\noindent \textbf{Geo-Object and Geo-Event.} \textit{Geo-Object} (e.g. a snapshot of the earth's atmosphere) and \textit{Geo-Event} (e.g. a hurricane) specialize the top-level ontological categories Endurant and Perdurant, respectively, from DOLCE. Both have a definite location in spatio-temporal space (represented as \textit{Spatio-Temporal Geo-Region}, a specialization of DOLCE's Spatio-Temporal-Region). The spatial and temporal footprints of this region is denoted by GeoSPARQL's Geometry and OWL-Time's TemporalEntity classes. \textit{Geo-Object} also specializes GeoSPARQL's class SpatialObject and is characterized by attributes (e.g. wind speed) and corresponding values (e.g. 130 knots).

\begin{figure}[h]
\caption{An RDF graph for events related to Hurricane Katrina modeled using the schema.}
\centering
\includegraphics[scale=0.5]{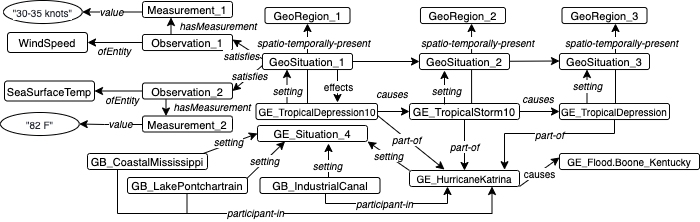}
\label{example}
\end{figure}

\noindent \textbf{Geo-Situation.} DOLCE's Situation (a non-physical endurant) specifies a state of the world for any instant of time and involves a set of constituents. \textit{Geo-Situation} in our proposed model extends this class and is associated with a set of observation parameters that correspond to attributes of the geo-objects situated therein. A geo-situation and the associated observations, which together represent a specific type of situation, are key to enriching causal discovery of geo-events. When a geo-situation satisfies one set of observation measurements (e.g. ocean temperature{\small $>82^{\circ}F$}, atmospheric pressure{\small  $>10^3 mb$}, wind shear{\small $>10 m/s$}, coriolis force {\small `\textit{present}'} is one set of conditions for a tropical cyclone), a geo-event is initiated, which creates a new situation temporally consequent to the situation that initiated the event. In that sense, geo-situations are temporally ordered, but they can also be nested as there can be a geo-event that has parts (e.g. see Fig~\ref{example}-Hurricane Katrina constituted a tropical wave, tropical storm, tropical depression, and several flood events).
Thus, the geographic world is a sequence of geo-situations, where each situation is defined by its physical setting -- geo-object(s) and/or geo-event(s) -- and any observations.
  
\noindent \textbf{General Relations.} Relations in the schema are inherited from OWL-Time, GeoSPARQL and DOLCE. Temporal parthood between a geo-event and its individual segments are denoted using DOLCE's \textit{part-of} relation. Every geo-object and geo-event is associated with a spatio-temporal geo-region via the \textit{spatio-temporally-present} relation. 
A geo-object evolves by participating in a geo-event through the \textit{participant-in} relation. Geo-object(s) and their participant geo-event both share the same spatio-temporal geo-region. This region is related to its spatial and temporal locations using the \textit{hasGeometry} and \textit{time} relations. Any geo-object or geo-event is situated in a geo-situation -- represented through the \textit{setting} relation, and each geo-situation \textit{satisfies} a set of observational measurements during the temporal duration that it holds.

\noindent \textbf{Causal and Causal-like relations.} Galton \cite{worboys2005event} makes the distinction between causal and 5 causal-like relations - where the two participating entities (events and states) are dependent on each other but neither necessarily causes the other. Here, we introduce three generic causal-relations -- \textit{causes}, \textit{effects} and \textit{affects} -- the first two are semantically similar in a general sense, but are distinguished in our model by their domain and range restrictions. A geo-event \textit{causes} another geo-event. A geo-situation that only constitutes \textit{geo-objects} and  whose observations satisfy a precondition \textit{effects} a geo-event. This distinction helps in capturing causality between sub-events when situation-specific information for asserting causality is not available. For example, a flood (geo-event) is either caused by heavy rains (geo-event) or effected by a dam whose water level is high and therefore overflowing (geo-situation), while the dam (geo-object) itself did not directly cause the flood. Geo-events are characterized by changes in the participating geo-objects and these changes can be traced by comparing the properties of these objects in different geo-situations, i.e. a geo-situation that is the setting for a geo-event \textit{affects} the participating geo-object(s). When a geo-situation leads up to a geo-event, then this situation holds for a temporal entity preceding the event itself. However due to space constraints we omit the formalization of the temporal connectedness between (geo)-events/situations, and partial order relation between geo-events, which requires modeling event partonomy. Fig~\ref{example} depicts a graph of a subset of events related to Hurricane Katrina using concepts and relations from the model.
\vspace{-0.2cm}

\vspace{-0.1cm}
\vspace{-0.3cm}
\section{Conclusion}
\vspace{-0.3cm}

This paper proposes a model to represent (geographic-) objects and events and their relationships with focus on event causality. The model introduces ``geographic situations" linked with observational snapshots as the linkage between each geographic event with a set of physical, chemical, or social factors that effect it, thereby helping answering why the event occurred. Causality in our model is defined in terms of a situation, i.e. what influences a geo-event to occur is an appropriate geo-situation. For example, the issue of how a tropical-depression event occured is actually a question of the temperature and wind attributes of a geo-situation. A geo-situation is the necessary condition for the effect. This causal-like relationship is justified by some (maybe implicit) underlying causal theory. Application-specific ontologies, e.g. a hurricane ontology, may extend the general schema by the application-specific observation conditions to help provide more fine-grained \textit{geo-situation}--\textit{geo-even}t causal relationships. We only mean to introduce the model’s basic concepts in this paper, and work towards a full presentation of the scheme and its axioms in the future.
\vspace{-0.2cm}


\vspace{-0.2cm}
\bibliography{lipics-v2019-sample-article}

\begin{thebibliography}{1}

\bibitem{devaraju2015formal}
Anusuriya Devaraju, Werner Kuhn, and Chris~S Renschler.
\newblock A formal model to infer geographic events from sensor observations.
\newblock {\em IJGIS}, 29(1):1--27, 2015.

\bibitem{galton2012states}
Antony Galton.
\newblock States, processes and events, and the ontology of causal relations.
\newblock In {\em Formal ontology in information systems}, pages 279--292. IOS
  Press, 2012.

\bibitem{grenon2004snap}
Pierre Grenon and Barry Smith.
\newblock Snap and span: Towards dynamic spatial ontology.
\newblock {\em Spatial cognition and computation}, 4(1):69--104, 2004.

\bibitem{llaves2014event}
Alejandro Llaves and Werner Kuhn.
\newblock An event abstraction layer for the integration of geosensor data.
\newblock {\em International Journal of Geographical Information Science},
  28(5):1085--1106, 2014.

\bibitem{scherp2012core}
Ansgar Scherp, Thomas Franz, Carsten Saathoff, and Steffen Staab.
\newblock A core ontology on events for representing occurrences in the real
  world.
\newblock {\em Multimedia Tools and Applications}, 2012.

\bibitem{shanahan1999event}
Murray Shanahan.
\newblock The event calculus explained.
\newblock In {\em Artificial intelligence today}. 1999.

\bibitem{stephen2017ontological}
Shirly Stephen and Torsten Hahmann.
\newblock An ontological framework for characterizing hydrological flow
  processes.
\newblock In {\em 13th International Conference on Spatial Information Theory},
  2017.

\bibitem{worboys2005event}
Michael Worboys.
\newblock Event-oriented approaches to geographic phenomena.
\newblock {\em IJGIS}, 2005.

\bibitem{zhang2005hierarchies}
Rui Zhang.
\newblock Hierarchies for event-based modeling of geographic phenomena.
\newblock 2005.

\end{thebibliography}

\end{document}